# Structured Occlusion Coding for Robust Face Recognition

Yandong Wen[a], Weiyang Liu[b], Meng Yang[c], Yuli Fu[a]*, Youjun Xiang[a], Rui Hu[a]

[a]*School of Electronic & Information Engineering, South China University of Technology, China.*
[b]*School of Electronic & Computer Engineering, Peking University, China.*
[c]*College of Computer Science & Software Engineering, Shenzhen University, China.*

**Abstract**

Occlusion in face recognition is a common yet challenging problem. While sparse representation based classification (SRC) has been shown promising performance in laboratory conditions (i.e. noiseless or random pixel corrupted), it performs much worse in practical scenarios. In this paper, we consider the practical face recognition problem, where the occlusions are predictable and available for sampling. We propose the structured occlusion coding (SOC) to address occlusion problems. The structured coding here lies in two folds. On one hand, we employ a structured dictionary for recognition. On the other hand, we propose to use the structured sparsity in this formulation. Specifically, SOC simultaneously separates the occlusion and classifies the image. In this way, the problem of recognizing an occluded image is turned into seeking a structured sparse solution on occlusion-appended dictionary. In order to construct a well-performing occlusion dictionary, we propose an occlusion mask estimating technique via locality constrained dictionary (LCD), showing striking improvement in occlusion sample. On a category-specific occlusion dictionary, we replace $l_1$ norm sparsity with the structured sparsity which is shown more robust, further enhancing the robustness of our approach. Moreover, SOC achieves significant improvement in handling large occlusion in real world. Extensive experiments are conducted on public data sets to validate the superiority of the proposed algorithm.

## 1. Introduction

Recent years have witnessed the rapid development of face recognition thanks to enormous theoretical breakthrough and increasingly strong computing capability. Nonetheless, most existing algorithms work fine under laboratory conditions but fail dramatically in practical, largely because of the variations they suffer in misalignment, illumination, expression and partial occlusion. Among all these problems, occlusion is viewed as the most common and challenging one. It is necessary for a practical face recognition system to handle occlusion, such as hats, scarves or sunglasses. Sometimes, exaggerated facial expression is also treated as another kind of occlusion. These occlusions may destroy essential discriminative information, leading to misclassification, especially when the occlusion occupies large region. In a practical face recognition system, the real-time ability is also necessary, since long-time waiting results in poor user experience and limits its applications.

In the literature of face recognition, traditional holistic feature based approaches [1, 2] are sensitive to outlier pixels, performing poorly against occlusion. Local feature based approaches are roughly divided into two categories in terms of patch-dependent [3-5] or data-dependent [6, 7]. Combined with partial matching methods [8, 9], the local features are more robust because they are not extracted from the entire image. Nonetheless, they are still inevitably affected by invalid pixels and far from being robust enough in practical classification tasks. An alternative solution addresses occlusion via a two-stage approach. It first identifies and discards the occluded pixels, and then performs the classification on the rest [10, 11]. As one can imagine, its classification performance is greatly determined by the

---

occlusion identification accuracy. If too much discriminative information is abandoned, the following classification becomes difficult. To enhance the accuracy of occlusion identification, [11] adopts the prior that the occlusion is spatially continuous and consequently achieves excellent performance. However, such unsupervised approach might cause misestimate when occlusion is severe. For instance, a scarf larger than half of the testing face may be considered as a useful signal, and therefore face pixels may be discarded in each iteration. We call it *a degenerate solution.* Besides, the algorithm in [11] has to be carried out subject-by-subject and exhaustively search the class with the minimum normalized error, which is time-consuming and detrimental to real-time applications.

Recently, several occlusion dictionary based approaches [13-16] for robust face recognition have been attached more and more importance. This kind of method is capable of efficiently handling various occlusions. They exploit characteristics of non-occluded and occluded region, assuming that both of them can be coded over the corresponding part of dictionary [12]. These methods act in the similar way with each other. Concretely, an occlusion dictionary is concatenated to the original dictionary to perform occlusion coding. The goal is to jointly represent the occluded image. Fig. 1 illustrates how occlusion dictionary methods work. By seeking a sparse solution, the occluded image successfully decomposes into face and occlusion. The classification is carried out via the corresponding coefficients. Hence, they cast the recognition problem with occlusion as the one without occlusion, since occlusion is regarded as an extra and special class in training samples. On the other hand, these occlusion dictionary based approaches choose different occlusion dictionary, leading to very different performance. More specifically, sparse representation-based classification (SRC) [13] employs an identity matrix as the occlusion dictionary, showing promising robustness to random pixel corruption and small contiguous occlusion. [14] exploits the local characteristics of Gabor feature, and proposes Gabor feature based sparse representation classification (GSRC). The Gabor feature of identity matrix exists high redundancy, so it can be compressed to a compact form for efficiency. Extended SRC (ESRC) [15] points out that intra-class variant dictionary contains useful information. By exploiting the difference images between two samples in the same class, ESRC can handle certain occlusion. The recent improvement is made by [16], namely structured sparse representation based classification (SSRC). They obtain common occlusion samples from occluded images with projection residuals, and utilize K-SVD [17] to train occlusion dictionary, making appended occlusion atoms to be more representative. SSRC achieves higher accuracy in both recovery and recognition.

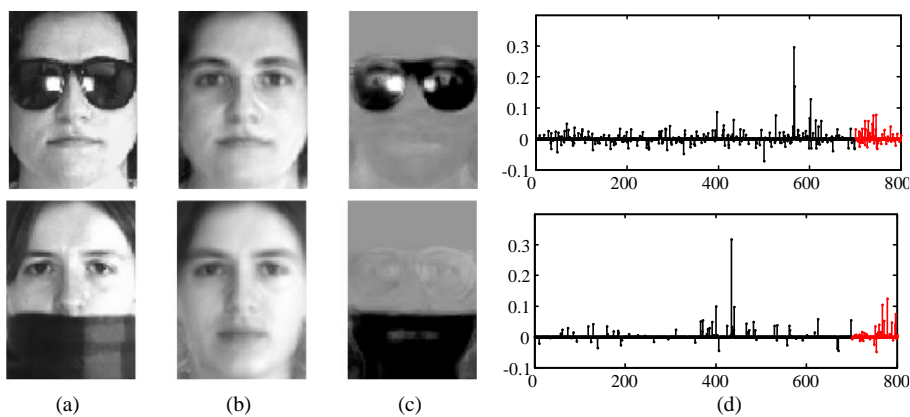

(a) (b) (c) (d)

Fig. 1 Overview of dictionary based approaches. An occluded image is coded over a compound dictionary under the $l_1$ sparse constraint. (a) The occluded test image. (b) The recovered face. (c) The recovered occlusion. (d) The sparse coefficients. Black stems refer to the coefficients of the original dictionary; red stems refer to the coefficients of the occlusion dictionary. Specifically, coefficients with indexes from 701 to 750 associate with *sunglasses atoms*, and coefficients with indexes from 750 to 800 associate with *scarf atoms.*

We consider a series of practical scenarios, e.g. office buildings, school or household surveillance, where strong robustness and excellent real-time efficiency are desired. In such scenarios, the occlusions are often limited and predictable. In addition, sufficient well-controlled training set including occluded and non-occluded training gallery images are available, following the same settings in [15, 16]. After slightly broadening the prior, we can take advantage of the training set to strengthen the robustness of our system when dealing with possible occlusions. Occlusion dictionary based methods can well address this kind of scenario. However, high dimensional test image forces the scale of identity matrix in [13] to be huge, which results in huge computational cost. Moreover, such a general occlusion dictionary [13, 14] can not distinguish between valid and invalid signal, performs poorly in separating face signal from occlusions. In [15], the atoms in intra-class variant dictionary have no explicit meaning to specific occlusion, suffering from severe inter-class interference. [16] collects the occlusion samples of a specific category, but extracts information of occlusion from projection residual of the whole image, which is not reasonable. Both [15] and [16] do not take the occlusion mask into account, leaving lots of non-occluded information in occlusion samples.

This paper aims to address the face recognition problem with occlusion in practical scenarios. We propose an efficient sparse representation based approach, namely structured occlusion coding (SOC), to simultaneously separate the occlusion and classify the face, via an occlusion-appended dictionary. In this case, face and occlusion can be represented respectively by corresponding part of the dictionary, leading to more effective and efficient classification in complex scenarios. By carefully designing the occlusion dictionary and properly harnessing the structured sparsity, our method is capable of handling large occlusion, like scarf, which sometimes covers more than 60% of the original face. In addition, our method can also deal with exaggerated facial expression, achieving significantly improvement. Interestingly, with such dictionary, not only the subject of face is well-classified, but the occlusion can also be identified. Moreover, invalid face and unknown occlusion tends to be rejected, via the proposed rejection rule.

The first contribution is that we propose an accurate and efficient approach to collect occlusion samples. Clearly, the occlusion dictionary is essential to the proposed algorithm but a well-trained occlusion dictionary is hard to design. First, the non-occluded information in occluded gallery images should be carefully excluded, which ensures as less interference as possible to the original dictionary for robust recognition. In order to obtain an exact occlusion sample, we use the occlusion mask to extract the pattern, so that we can ignore the non-occluded information. In this case, the technique of estimating occlusion mask is utilized in advance. The pioneering work in [11] is a nice choice yet still with a major limitation that the label of occluded image and the corresponding training samples are required. In this paper, we propose to collect occlusion samples via locality constrained dictionary (LCD) [18, 19]. Such approach does not need the label of occluded image. Our method achieves comparable performance to [11], while can be used in a wider range of scenarios. Second, a compact occlusion dictionary is the key to efficiency. We should not include many redundant atoms in the occlusion dictionary. Therefore, K-SVD [17] algorithm is employed to learn a dictionary from the occlusion samples.

Another significant contribution is the structured sparsity [20] we introduce in the proposed method for stronger robustness. Since this paper aims at recognition rather than representation, a structured sparse solution provides more discrimination and robustness, decreasing the interference between unrelated face and occlusion atoms. Fig. 5 shows an example of structured sparsity penalty on solution over a compound dictionary. Because of the structured sparsity constraint, the original rejection mechanism in SRC [13], namely sparsity concentrate index (SCI), is no longer applicable. Moreover, in the case that both the coefficients of valid and invalid images are not sparse, the residuals still keep their discriminative ability as shown in Fig. 6. We naturally propose a more generic rule based on residuals, namely residual distribution index (RDI) to reject an invalid face, even unknown occlusion, whose superiority is validated in experiments.

The rest of this paper is organized as follow. Section 2 briefly reviews SRC. We propose the structured sparse coding with an occlusion dictionary for robust face recognition in section 3, including the various details in classifying and rejecting. Comprehensive experimental results are presented and analyzed in section 4, demonstrating the superiority of our algorithm. Last, we conclude this paper in section 5.

## 2. Sparse representation-based classification

Given an $a \times b$ frontal face image, we only consider grayscale case. Each face image is stacked into a vector $\boldsymbol{d}_{i,1} \in \mathbb{R}^m (m = a \times b)$ which denotes the $j$th samples from $i$th class. Constituting all vectors of the $i$th subject gives a sub-dictionary $D_i = [\boldsymbol{d}_{i,1}, \boldsymbol{d}_{i,2}, ..., \boldsymbol{d}_{i,n_i}] \in \mathbb{R}^{m \times n_i}$. We assume that training samples belong to $k$ distinct classes. $k$ sub-dictionaries are put together as a dictionary matrix with the form of $D = [D_1, D_2, ..., D_k] \in \mathbb{R}^{m \times n}$. According to the low-dimensional linear illumination model [12], a non-occluded testing sample $y \in \mathbb{R}^m$ belonging to the $r$th class can be well represented as a linear combination of atoms in $D$ . i.e. $y = \sum_{i=1}^{k} \sum_{j=1}^{n_i} x_{i,j} \boldsymbol{d}_{i,i} = \sum_{i=1}^{k} D_i \boldsymbol{x}_i$ , where $\boldsymbol{x}_i \in \mathbb{R}^{n_i \times 1}$ and $\boldsymbol{x}_i = [x_{i,1}, x_{i,2}, ..., x_{i,n_i}]^T$ are the coefficients corresponding to the $i$th class. SRC [13] states that in most cases, the linear combination of those training sample from the $r$th class is the sparest representation which can be solved via $l_0$-minimization.

$$\hat{\boldsymbol{x}}_0 = \arg\min_{\boldsymbol{x}} \|\boldsymbol{x}\|_0 \quad \text{s.t.} \quad D\boldsymbol{x} = \boldsymbol{y} \tag{1}$$

In the ideal case, the coefficient vector should be $\boldsymbol{x} = [0, ..., 0, \boldsymbol{x}_{r,1}, \boldsymbol{x}_{r,2}, ..., \boldsymbol{x}_{r,n_r}, 0, ..., 0]^T \in \mathbb{R}^n$ whose entries are zero except those associated with the $r$th class. Such sparse coefficients $\boldsymbol{x}$ contains the discriminative information, revealing the label of $\boldsymbol{y}$. The classification result is given by the class with minimal residual. Denoted by $\delta_i(\boldsymbol{x}) = [0, ..., 0, \boldsymbol{x}_{i,1}, \boldsymbol{x}_{i,2}, ..., \boldsymbol{x}_{i,n_i}, 0, ..., 0]^T$, the coefficient vector for class $i$ is preserved while the coefficients of the other classes are set to zero. The label of the testing sample $\boldsymbol{y}$ is given by

$$\min_{i} r_i(\boldsymbol{y}) = \|\boldsymbol{y} - D\delta_i(\boldsymbol{x})\|_2 \tag{2}$$

## 3. The proposed algorithm

In this section, we elaborate the proposed framework in which we cast both occlusion separation and classification as a problem of seeking a structured sparse representation over a compound dictionary. We first formulate the classification problem with practical occlusion, followed by the classification and rejection mechanism to face and occlusion. Next, we illustrate how we estimate the occlusion mask and extract its pattern. After making a comparison with several widely used techniques of collecting occlusion samples, we present the proposed algorithm, and show that it achieves competitive performance without label information. Last, we discuss the difference between $l_1$ sparse and structured sparse, arguing that structured sparse can better address the recognition task with occlusion dictionary.

*3.1. The classification framework*

Assuming a face image $\boldsymbol{y}$ is occluded by an object $\boldsymbol{v}$ as the form of $\boldsymbol{u} = \boldsymbol{y} + \boldsymbol{v}$ [16], the low-dimension linear illumination model is violated, resulting in misclassification of SRC. As mentioned above, the categories of occlusions in real scenarios are predictable and can be collected in advance, which inspires us to concatenate $s$ occlusion sub-dictionaries $B = [B_1, B_2, ..., B_s]$ to the original dictionary, forming a compound dictionary $R = [D, B]$. Note that different subscripts indicate different categories of occlusion. Similar to faces, a specific occlusion $\boldsymbol{v}$ belonging to the $t$th class can be well represented by the corresponding sub-dictionary $B_t$, which should be the sparest representation over the dictionary as well [13]. Specifically, if a face of the $r$th subject is occluded by an occlusion of the $t$th category, then $\boldsymbol{u}$ has its linear representation of by the atoms from $D_r$ and $B_t$, with the form of $\boldsymbol{u} = \boldsymbol{y} + \boldsymbol{v} = D_r\boldsymbol{x}_r + B_t\boldsymbol{c}_t$. The original linear illumination model is maintained with the occlusion appended. By seeking proper sparse solution, we are able to obtain a series of coefficients whose non-zero entries reveal the subject of face and occlusion. Concretely, the ideal solution should be the form of $\boldsymbol{w} = [\boldsymbol{x}^T, \boldsymbol{c}^T]^T = [0, ..., 0, \boldsymbol{x}_{r,1}, \boldsymbol{x}_{r,2}, ..., \boldsymbol{x}_{r,n_r}, 0, ..., 0, \boldsymbol{c}_{t,1}, \boldsymbol{c}_{t,2}, ..., \boldsymbol{c}_{t,n_t}, 0, ..., 0]^T$. Therefore, a well-constructed dictionary is able to handle various occlusion, being highly robust in practical scenario. The recognition problem with occlusion is formulated as

$$\hat{\boldsymbol{w}}_0 = \arg\min_{\boldsymbol{w}} \|\boldsymbol{w}\|_0 = \arg\min_{\boldsymbol{x},\boldsymbol{c}} \|\boldsymbol{x}; \boldsymbol{c}\|_0 \quad \text{s.t.} \quad \left\|\boldsymbol{u} - [D, B]\begin{bmatrix}\boldsymbol{x}\\\boldsymbol{c}\end{bmatrix}\right\|_2 = \|\boldsymbol{u} - R\boldsymbol{w}\|_2 \leq \varepsilon. \tag{3}$$

By solving problem (3), the label assignment of the face can be operated on $\hat{\boldsymbol{x}}$ and $D$. Since finding sparest solution of equation (3) is NP-hard [21], it is impractical to solve this problem as the number of atoms increases. Thanks to theoretical breakthrough in compressed sensing [22-24], which reveals that $l_0$-minimization problem (3) is equivalent to $l_1$-minimization problem under certain condition. Thus the minimization Eq. (3) becomes

$$\hat{\boldsymbol{w}}_1 = \arg\min_{\boldsymbol{w}} \|\boldsymbol{w}\|_1 = \arg\min_{\boldsymbol{x},\boldsymbol{c}} \|\boldsymbol{x}; \boldsymbol{c}\|_1 \quad \text{s.t.} \quad \left\|\boldsymbol{u} - [D, B]\begin{bmatrix}\boldsymbol{x}\\\boldsymbol{c}\end{bmatrix}\right\|_2 = \|\boldsymbol{u} - R\boldsymbol{w}\|_2 \leq \varepsilon. \tag{4}$$

We calculate the residual between $\hat{\boldsymbol{y}}$ and $\hat{\boldsymbol{y}}_i$, where $\hat{\boldsymbol{y}}_i = D\delta_i(\hat{\boldsymbol{x}})$. In spite of occlusion, we still enable the test sample $u$ to be assigned to the subject with minimal residual. Due to the special structure of the compound dictionary, the category of occlusion can still be identified via the coefficients spanning on the occlusion dictionary $B$. We summarize both classification rules as follows:

$$\min_i r_i(\bm{y}) = \|\bm{u} - \hat{\bm{y}}_i - B\hat{\bm{c}}\|_2$$
$$\min_i r_i(\bm{v}) = \|\bm{u} - \hat{\bm{v}}_i - D\hat{\bm{x}}\|_2. \qquad (5)$$

*3.2. Occlusion pattern acquisition from occluded face images*

Recently Zhou et al [11] proposes a novel technique to deal with continuous occlusion. They iteratively estimate occluded pixels by sparse error and graph model, and eventually reveal the occlusion mask of test image. Since we can easily get access to the identity of occluded image in training stage, it is unnecessary to exhaustively compute on each sub-dictionary to find out the best estimate, which however is unavoidable in [11]. By carefully choosing parameters, it stably and precisely estimates the occlusion region, or occlusion mask over the corresponding sub-dictionary, which enables us to capture an exact occlusion pattern from the occluded images. We show the estimate results in each iteration in Fig. 2. With the estimated mask, we can simply ignore the non-occluded part and collect the occlusion pattern by calculating the residual only in occluded region.

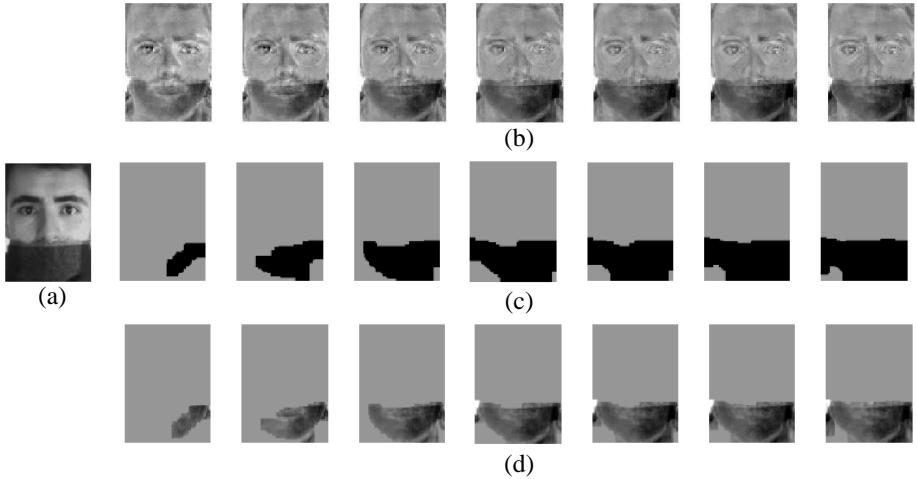

Fig. 2 The occlusion estimate results of an occluded image. (a) The occluded image (scarf). (b) Estimated error via locality constrained dictionary in each iteration. (c) Estimated occlusion mask in each iteration. Gray and black regions stand for non-occluded and occluded part of test image, respectively. (d) Estimated occlusion pattern after taking occlusion mask into account.

Since the label of occluded images is given a prior, we can easily estimate occlusion pattern via its associated sub-dictionary. In practical conditions, however, labels of some occluded images may be unknown, so that the algorithm in [11] is inapplicable. On the other hand, it is more likely to collect a vast variety of unlabeled occluded images rather than labeled ones. So how can we extract the occlusion pattern of the occluded images whose labels are unknown?

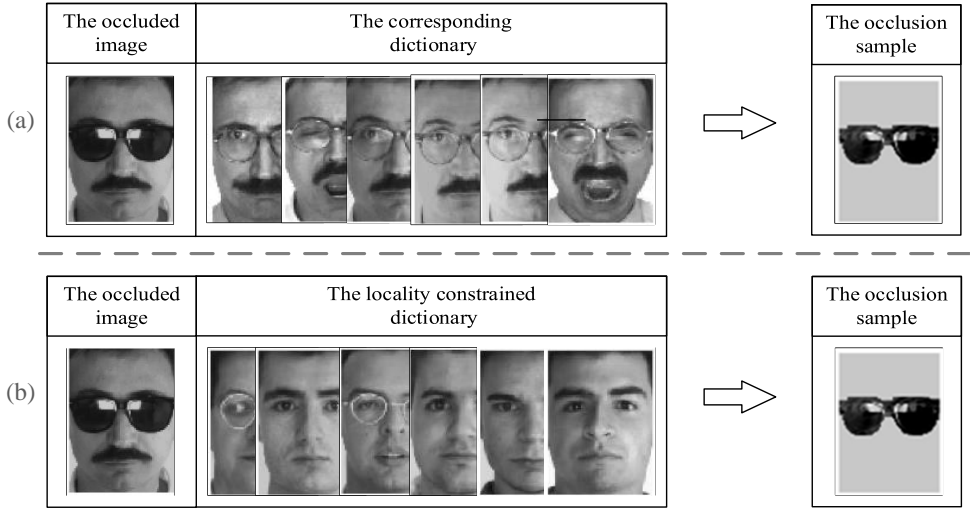

Fig. 3 (a) Estimate occlusion pattern by the corresponding dictionary. (b) Estimate occlusion pattern by locally constrained dictionary.

Inspired by LCD [18, 19], we propose an improved algorithm to deal with this problem. That is, estimating the occlusion region of the unlabeled test images with pre-existing original dictionary. We claims that the LCD provides a more generic and flexible approach to construct a sub-dictionary for estimating occlusion mask, working well without the essential prior knowledge. The proposed algorithm is given .

---

**Algorithm 1**: Occlusion Region Estimation via Locality Constrained Dictionary

1: **Input:** A matrix of normalized training samples $D \in \mathbb{R}^{m \times n}$, a test sample $\boldsymbol{u} \in \mathbb{R}^m$ with unknown label.
2: Measure the similarities between testing and each training sample by $\boldsymbol{\psi} = D^T \cdot \boldsymbol{u}$, $\boldsymbol{\psi} \in \mathbb{R}^n$ ($\boldsymbol{\psi}$ is the inner product of $D$ and $\boldsymbol{u}$) and choose the atoms with largest $h$ values in $\boldsymbol{\psi}$ to construct locality constrained dictionary $D_{LCD}$.
3: Initialize the error support $\boldsymbol{z}^{(0)} = 1_m$, $t = 0$.
4: **repeat**
5: $t = t + 1$;
5: $D^*_{LCD} = D_{LCD}[\boldsymbol{z}^{(t-1)} = 1, :], \boldsymbol{u}^* = \boldsymbol{u}[\boldsymbol{z}^{(t-1)} = 1]$;
6: Solve the convex program
$$(\hat{\boldsymbol{x}}, \hat{\boldsymbol{e}}^*) = \arg \min_{\boldsymbol{e}^*} \|\boldsymbol{e}^*\|_1$$
$$\text{s.t.} \quad \boldsymbol{u}^* = D^*_{LCD} \boldsymbol{x} + \boldsymbol{e}^*$$
7: $\hat{\boldsymbol{e}} = \boldsymbol{u} - D_{LCD} \hat{\boldsymbol{x}}$;
8: Update error support via graph cuts:
$$\boldsymbol{z}^{(t)} = \arg \max_{\boldsymbol{z} \in \{0,1\}^m} \sum_{i,j \in E} \beta \boldsymbol{z}[i]\boldsymbol{z}[j] + \sum_{i \in V} \log p(\hat{\boldsymbol{e}}[i]|\boldsymbol{z}[i]);$$
9: **Until** maximum iterations or convergence.
10: $\hat{\boldsymbol{e}}[\boldsymbol{z}^{(t)} = 1] = 0$;
11: **Output**: $\hat{e}$

---

In Algorithm 1, we consider the image domain as a graph $G = (V, E)$. $V = \{1, 2, ..., m\}$ denotes the indexes of $m$ pixels and $E$ denotes the set of edges connecting neighboring pixels. The support vector of error $\boldsymbol{e}$ is indicated as $\boldsymbol{z} \in \{0, 1\}^m$, where $\boldsymbol{z}[i] = 1$ or $\boldsymbol{z}[i] = 0$ means the $i$th pixel is regarded as non-occluded or occluded, respectively. The parameter $\beta$ controls the degree of mutual interaction between adjacent pixels. The log-likelihood function $\log p(\boldsymbol{e}[i]|\boldsymbol{z}[i])$ is given in (6). More details can refer to [11].

$$\begin{array}{rcl}
\log p(\boldsymbol{e}[i]|\boldsymbol{z}[i]=1) & = & \begin{cases} -\log \tau & \text{if } |\boldsymbol{e}[i]| \leq \tau, \\ \log \tau & \text{if } |\boldsymbol{e}[i]| > \tau, \end{cases} \\
\log p(\boldsymbol{e}[i]|\boldsymbol{z}[i]=0) & = & \begin{cases} 0 & \text{if } |\boldsymbol{e}[i]| > \tau, \\ \log \tau & \text{if } |\boldsymbol{e}[i]| \leq \tau, \end{cases}
\end{array} \quad (6)$$

Compared with the occlusion samples in [15] and [16], our algorithm removes as much noise as possible by incorporating the occlusion mask. As shown in Fig. 4, we obtain more exact occlusion pattern, which decreases the interference with face atoms and contributes to the subsequent classification.

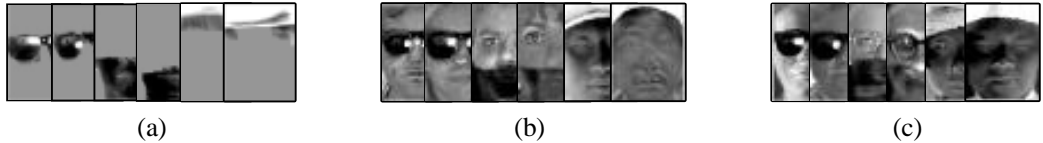

(a)          (b)          (c)

Fig. 4 (a) Examples of occlusion samples in SOC. (b) Examples of occlusion samples in SSRC[16]. (c) Examples of occlusion samples in ESRC [15].

*3.3. Structured sparsity for robust recognition*

Usually, it is inevitable for these sub-dictionaries to share a high degree of coherence. In a compound dictionary, the linear combination of the face component to represent testing images is possible to contain atoms from irrelevant classes, resulting in misclassification. In Fig. 2(d), a few non-zero entries occur in *scarf atoms*, since sparse representation does not take the dictionary structure into account. It is the individual atom in the candidate list, which really matters in representation. While the structured sparsity significantly avoids the interference from irrelevant sub-dictionaries, as shown in Fig. 5(d), negligible coefficients on *scarf atoms* are recovered under the structured sparsity constraint, which benefits the subsequent classification.

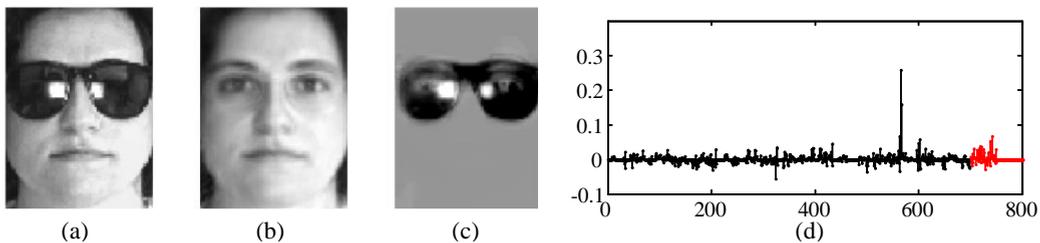

(a)          (b)          (c)          (d)

Fig. 5 An occluded image is coded under the structured sparsity constraint. (a) Test image. (b) The recovered face. (c) The recovered occlusion. (d) The sparse representation coefficients, half of the red stems associated with scarf atoms are zero, avoiding the interference from sunglasses.

Structured sparsity is proposed in [20] for robust recognition tasks, but [20] can not achieve satisfactory performance when dealing with occlusions. Specifically, it maintains the identity matrix as occlusion dictionary, which is a generalized one and does not belong to any specific category of occlusion. So a suboptimal strategy is to treat each column of the identity matrix as a block of length 1. In our category-specific occlusion dictionary, structured sparsity is smoothly combined for stronger robustness. With respect to the intrinsic structure of the compound dictionary, structured sparsity better reveals the potential information in occluded images. We penalizes the coefficients with structured sparsity constraint as follows:

$$\hat{\boldsymbol{w}}_0 = \arg\min_{\boldsymbol{w}} \sum_{i=1}^{k} I\left(\|\boldsymbol{x}_i\|_q > 0\right) + \lambda \sum_{i=1}^{s} I\left(\|\boldsymbol{c}_i\|_q > 0\right) \quad \text{s.t.} \quad \left\|\boldsymbol{u} - [D, B]\begin{bmatrix} \boldsymbol{x} \\ \boldsymbol{c} \end{bmatrix}\right\|_2 = \|\boldsymbol{u} - R\boldsymbol{w}\|_2 \leq \varepsilon \quad (7)$$

where $q \geq 1$ and $I(\cdot)$ is the indicator function, which equals to zero when it argument is zero and is one otherwise. The alternative formulation is give in (8), under certain condition [20], (8) has the same solution with problem (7).

$$\hat{\boldsymbol{w}}_1 = \arg\min_{\boldsymbol{w}} \sum_{i=1}^{k} \|\boldsymbol{x}_i\|_q + \lambda \sum_{i=1}^{s} \|\boldsymbol{c}_i\|_q \quad \text{s.t.} \quad \left\|\boldsymbol{u} - [D, B]\begin{bmatrix}\boldsymbol{x}\\\boldsymbol{c}\end{bmatrix}\right\|_2 = \|\boldsymbol{u} - R\boldsymbol{w}\|_2 \leq \varepsilon \qquad (8)$$

Similarly, we apply Eq. (5) in the subsequent decision making. However, a practical face recognition system may confront with sophisticated scenarios, such as occluded image, invalid image (subjects that are not included in original dictionary) or unknown occlusion (category not included in occlusion dictionary). Sparsity concentrate index (SCI), originally proposed in [13], works well in most cases but fails when both the valid and invalid test images cannot be sparsely coded, e.g. the feature dimension of test image is high. Besides, since the optimization problem is not penalized by $l_1$ sparsity constraint, it is not feasible to reject invalid images via SCI.

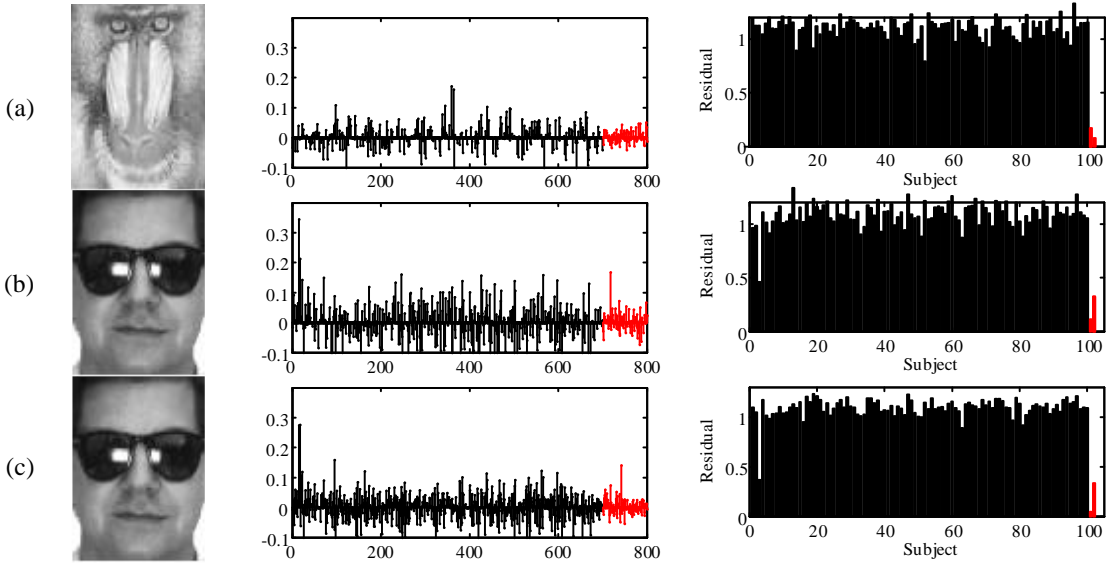

Fig. 6 (a) An invalid image and its corresponding coefficients and residuals. (b) A valid image and its corresponding $l_1$ sparse coefficients and residuals (c) A valid image and its structured sparse coefficients and residuals. When the coefficients are not sparse, the validity still can be recognized by residual distribution.

In this paper, we employ a residual-based rule, namely RDI, to verify the validity of the test image, which is shown effective in our experiments. A valid test image has the distinct difference in residual distribution between the related and unrelated class. In contrast, uniform distribution among all classes indicates that the test image does not belong to any subject in the training dictionary. Similarly, the rejection rule is also applied to occlusion. An unknown occlusion usually produces nearly equal residuals. We give the rejection rules of faces and occlusions in equation (9). Note that, if $k$ or $s$ is equal to 1, it is no longer a classification task. We reject a test image or an occlusion if $\text{RDI} > \theta$. RDI is defined as

$$\text{RDI}(\boldsymbol{y}) = \frac{k \cdot \min r_i(\boldsymbol{y})}{\sum_{i=1}^{k} r_i(\boldsymbol{y})}$$

$$\text{RDI}(\boldsymbol{v}) = \frac{s \cdot \min r_i(\boldsymbol{v})}{\sum_{i=1}^{s} r_i(\boldsymbol{v})}. \qquad (9)$$

The whole framework is shown in Fig. 7. In training stage, we collect occlusion samples from various occluded images, with or without labels. Next, we learn a compact occlusion dictionary from occlusion sample and append it to the original dictionary. In testing stage, given a test image, we solve the structured sparse problem over the compound dictionary, and perform classification on face and occlusion respectively.

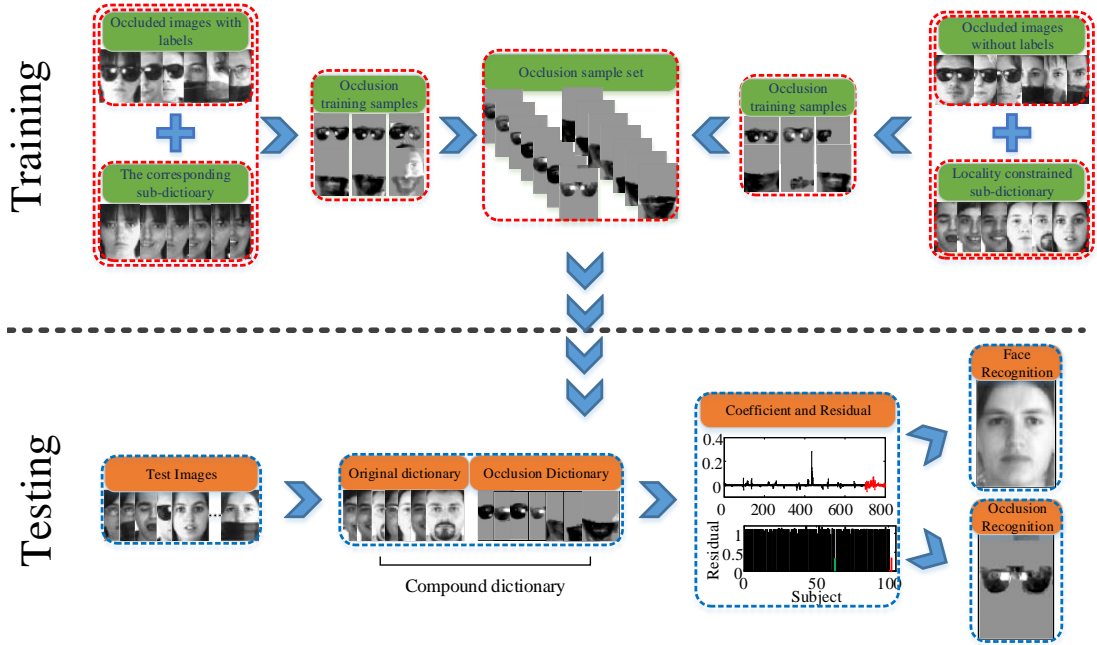

Fig. 7 The entire testing and training framework of SOC.

## 4. Experimental results

In this section, we conduct a series of experiments to validate our proposed algorithm. We compare SOC to most existing occlusion dictionary based methods, including SRC[13], GSRC[14], ESRC[15], SSRC[16]. SRC handles occlusion or random corrupted pixels with an identity matrix, which needs no extra computation in collecting samples and training dictionary. GSRC extracts Gabor local feature for more robustness, and promotes efficiency by reducing redundancy in appended dictionary. ESRC deals with occlusion by constructing an intra-class variant dictionary. SSRC, however, explicitly extracts specific occlusion pattern, and trains the obtained samples for better representation and less interference with original dictionary. In the experiments, we test the robustness of SOC to various occlusion in real world. Moreover, we state that exaggerated facial expressions can also be treated as a generalized occlusion, being addressed properly in SOC. We respectively solve $l_1$ sparse representation problem (4) and structured sparse representation problem (8), and present both recognition result, to further imply that structured sparse is more suitable for classification task.

We begin with the implementation details of experiments in section 4.1, and then discuss different methods to collect occlusion samples in section 4.2. Section 4.3 compare the performance of SOC, with four occlusion dictionary based methods in real face occlusion on AR [25] and CAS-PEAL [26] databases. Note that the occluded training samples in section 4.3 are labeled. While in section 4.4, we investigate the performance of SOC with unlabeled occluded training images, which is a more restrict environment. Next, we demonstrate the robustness of SOC to various expression, giving in section 4.5. Section 4.6 compares the running times of all the algorithms we mentioned, and briefly analyses from the view of practical application. Moreover, we state that not only the face can be recognized, the category of occlusion is also capable to be identified with our occlusion dictionary, which is showed in section 4.7. Finally, the ability to reject invalid images and occlusion (subjects or occlusion not present in the dictionary) is tested in section 4.8, and section 4.9 exploits the relationship between the recognition rate and the size of occlusion of dictionary.

*4.1. Implementation details*

In training stage, we obtain occlusion samples from the images with the resolution of $83 \times 60$. The parameter $\tau$ in algorithm 1 is choose adaptively, varying from 0.005 to 0.002 with a constant step -0.0005. In experiment 4.3 and 4.4, the parameter $\beta$ is set to a fixed value of 20, while decreases to 5 for expression extracting in 4.5. The size $h$ of locality constrained dictionary set to 20. The implementation of $l_1$-minimization is Primal Augmented Lagrangian Method (PALM)[27], which is an accelerated convex optimization techniques, with the complexity $O(n^2)$. It means that the running time of PALM only related to the number of atoms. Besides, PALM is also employed to solve the $l_1$-minimization problem in KSVD for efficient occlusion dictionary learning.

In test stage, we used CVX [28, 29] to solve problem (4) and (8) in the following experiments for fair comparison. The reconstruction error $\varepsilon$ is set to 0.05 in all the experiments. In formulation (8), $q$ is 2 and the parameter $\lambda$ is set to a constant $\sqrt{n_i/n_j}$, according to the amount of atoms in occlusion and face sub-dictionary. All the experiments are repeated 10 times and we calculate the average value for accuracy, performing in MATLAB with 3.60GHz Intel Core i5 processor and 8G memory.

*4.2. Occlusion dictionary construction*

In section 4.3, we utilize algorithm in [11] to exactly estimate occlusion sample with $D_r$ from $\boldsymbol{u}_r$, where both the occluded training image $\boldsymbol{u}_r$ and sub-dictionary $D_r$ belong to the $r$th class. The approach of obtaining occlusion sample in SSRC formulates as $\boldsymbol{u}_r - D_r(D_r^T D_r)^{-1} D_r^T \boldsymbol{u}_r$[16]. For ESRC, occlusion sample are obtained by subtracting the occluded image from the centroid of sub-dictionary $D_r$, giving as $\boldsymbol{u}_r - 1/n_i \sum_{j=1}^{n_i} \boldsymbol{d}_{i,j}$. The Gabor occlusion dictionary and Gabor feature are calculated using the code published by authors[14].

In a more restrict scenario (section 4.4), the label of test image is not available, so we utilize algorithm 1 in occlusion samples collecting. From the entire dictionary $D$, we compute the locality constrained dictionary $D_{LCD}$ and estimate the occlusion from $\boldsymbol{u}$. On the other hand, SSRC replace the $D_r$ by a arbitrary sub-dictionary and ESRC chooses to subtract the test image from the centroid of dictionary $D$.

Intuitively, we can simply append all or some of occlusion samples to original dictionary, which also a feasible strategy to construct the occlusion dictionary. Due to the high redundancy in occlusion samples, it is possible to compress them for efficiency of test stage. Fig. 8 shows the eigenvalues of one of the occlusion sub-dictionary, giving a clue that the occlusion sample set is redundant and compressible.

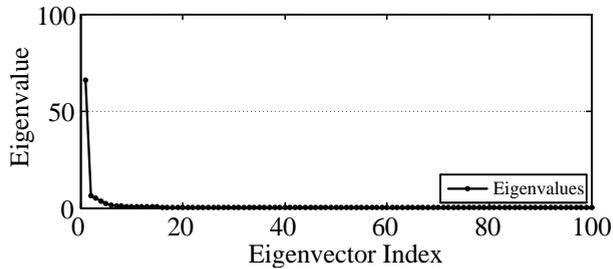

Fig. 8 eigenvalues of occlusion samples (sunglasses)

There is a rich literature with respect to this classical task of learning a compact and representative dictionary from a sample set. Liu [30] incorporate clustering techniques in dictionary construction, significantly compressing number of atoms so that lower computational cost can be achieved. Some dictionary learning algorithm are proved to be effective to this task [31], successfully training the dictionary to be a compact one without losing classification accuracy. For simplify, we employ K-SVD [17] technique in dictionary training. Fig. 9 shows how we learn an occlusion dictionary from occlusion samples. We can know that the occlusion dictionary in SSRC and ESRC are noisy due to the remaining non-occluded information in occlusion samples.

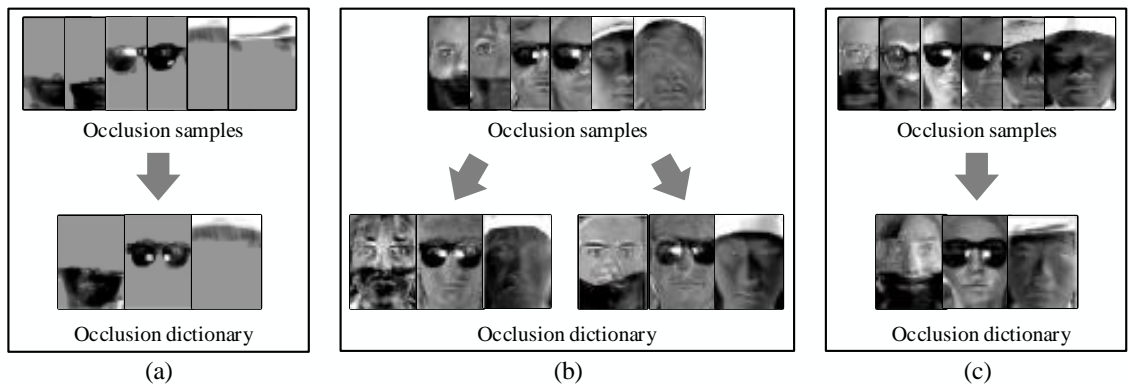

Fig.9 Different types of occlusion samples extracted by three kind of approaches, and we learn an occlusion dictionary from occlusion samples. (a) SOC (b) SSRC1 and SSRC2 (c) ESRC.

*4.3. Robust face recognition with labeled occluded images*

We conduct the experiments on the AR database and CAS-PEAL database. In this scenario, sufficient non-occluded images are provided to construct the original dictionary, and occluded images with labels are available for occlusion dictionary training. Our task is to carry out classification on non-occluded or occluded testing face images.

*4.3.1 The AR database*

The AR database collects over 4,000 frontal images for 126 individuals. These images include more facial variations, including illumination change, expressions, and facial disguises. We choose a subset of 50 men and 50 women with 26 images (14 normal images, 6 images with sunglasses and 6 images with scarf) each subject. For each subject, 7 images without any disguises are randomly selected to construct the original dictionary and the others 7 images use for testing. In sunglasses and scarf gallery images, we randomly pick 10 subjects (60 images) respectively to obtain occlusion samples by the approaches elaborated in 4.2. The rest of images are put into testing set. The amount of occlusion atom is 30 in GSRC, ESRC, SSRC1, SSRC2 and SOC for fair comparison. We compute the recognition ratio with the down-sampling resolution of $12 \times 10$, $14 \times 12$, $17 \times 15$, $21 \times 20$ and $28 \times 20$ associated with the feature dimensions 120D, 168D, 255D, 420D and 560D.

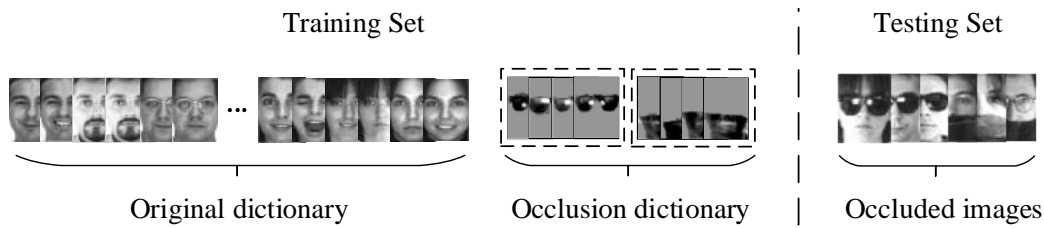

Fig. 10 Diagram of face recognition task with occlusion (sunglasses and scarf) on AR database.

**Scenario 1:** We first test SOC in the single occlusion scenario. The occlusion dictionary of 30 atoms is learned from 60 images with sunglasses, together with original dictionary to form the training set. The remaining 540 images with sunglasses are used for testing. The results are given in Fig. 11(a).

**Scenario 2:** Another single occlusion scenario is presented. The original dictionary is kept unchanged but the category of occlusion dictionary are replaced by scarf, sharing the same amount of atoms and training samples. Certainly, the testing set consists of 540 images with scarf. It is relatively difficult than scenario 1, since the occlusion area is larger. Fig. 11(b) shows the recognition rates.

**Scenario 3:** In this scenario, we consider the multiple occlusions scenario. It is the most challenging one due to the unknown category of occlusion and mutual interference among sub-dictionaries. The training set includes original dictionary, occlusion dictionary of sunglasses and scarf. The rest of 1080 occluded images are used for testing. Fig. 11(c) gives the experimental results.

**Scenario 4:** Classification performance on normal images are also presented because a practical face recognition system should be able to both handle occluded and non-occluded testing images. We choose the same training set with scenario 3. The rest of 699 images without disguise (a corrupted image w-027-14.bmp is discarded) are used for testing. The results are shown in Fig. 11 (d).

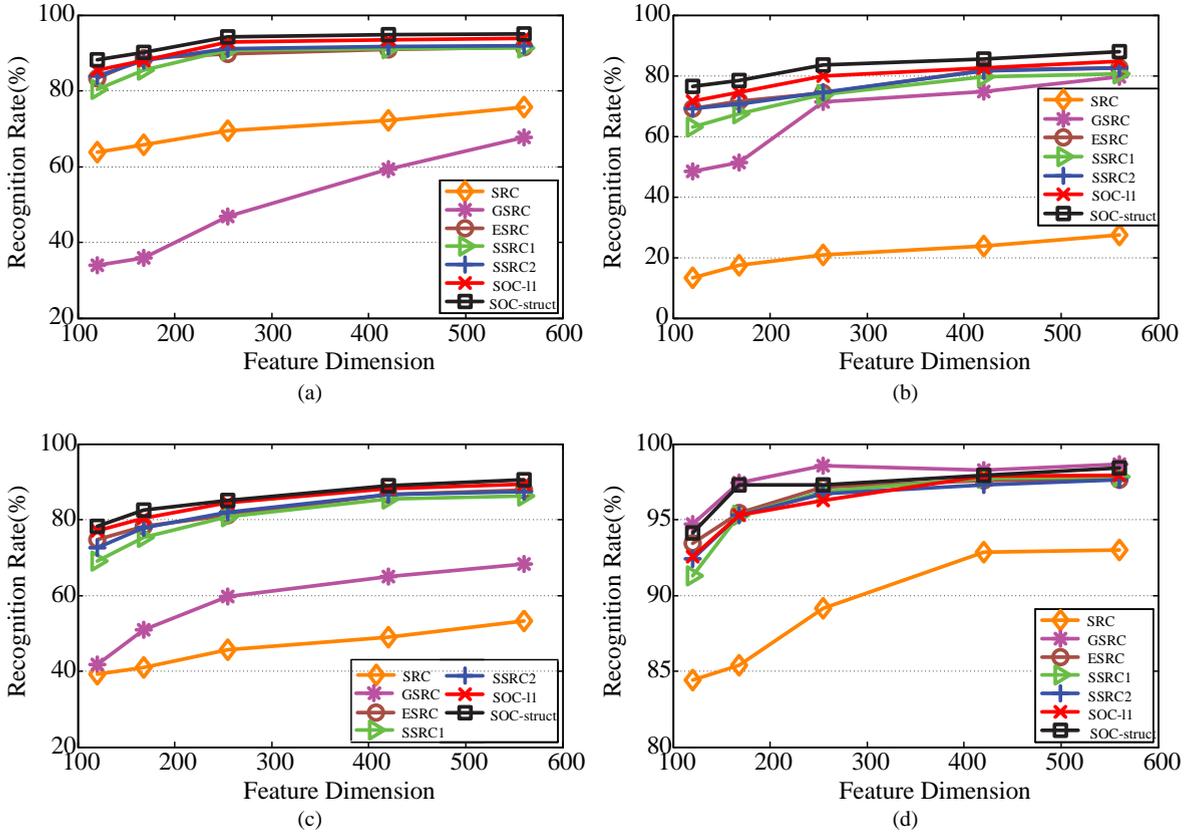

Fig. 11 The recognition rate in four scenarios on AR database. (a) Sunglasses images. (b) Scarf images. (c) Sunglasses and scarf images (d) Normal images

It is obvious that SRC achieves poor recognition rate in large occlusion (scarf), while performs much better when less area is occluded (sunglasses, normal images). It means that SRC is sensitive to the amount of occluded pixels. The main reason is that large occlusion results in dense solution, violating the precondition of $l_1$-minimization. In contrast, GSRC is relatively stable to different occlusion, and achieves best performance in normal face scenario. When image is occluded, the performance of GSRC decreases rapidly in low dimension. Similar with SRC, such a generic occlusion dictionary, it can not handles the specific occlusion as well as category-specific occlusion dictionary based methods. ESRC, SSRC1 and SSRC2 address in specific occlusion, applying KSVD to learn an occlusion dictionary from samples. Both the difference in image domain and projection residual provide sufficient information about occlusion, which is benefit to represent the occluded test image. So it outperforms the SRC, GSRC.

Nonetheless, the strategy of collecting occlusion samples, from the perspective of a whole image, inevitably detrimental to representation and classification. Incorporating occlusion mask in occlusion pattern collecting removes most non-occluded information, greatly enhancing the generalization of occlusion dictionary. By carefully constructing the occlusion dictionary, we significantly improves the recognition accuracy, respectively achieving the highest recognition rate at 95.07%, 87.92%, 90.47% and 98.40% in the four scenarios. Besides, our algorithm outperforms SSRC1, especially in low dimension condition (120D), with 7.78%, 13.33%, 9.26% and 2.86% in four scenarios, respectively. It is also worth mentioning that our algorithm performs better with structured sparsity than traditional $l_1$ sparsity constraint, especially in 120D situation. The results improvement of 2.78%, 4.81%, 1.29% and 1.57% are presented.

*4.3.2 The CAS-PEAL database*

The CAS-PEAL face database contains 99,594 images of 1040 individuals (595 males and 445 females) with varying Pose, Expression, Accessory, and Lighting. It is worth of mentioning that this database is substantially more challenging, since the hat leaves a shadow area in forehead and eyes, covering lots of useful information. We choose a subset of 100 subjects. Each subject contains 10 images under various illumination and 3 images with hat. Again, we randomly pick 10 subjects (30 images) for training and the others for testing. We set the size of occlusion dictionary to 20, so we have 1020 atoms in dictionary and 270 images in test set. The experimental results are shown in Fig. 13.

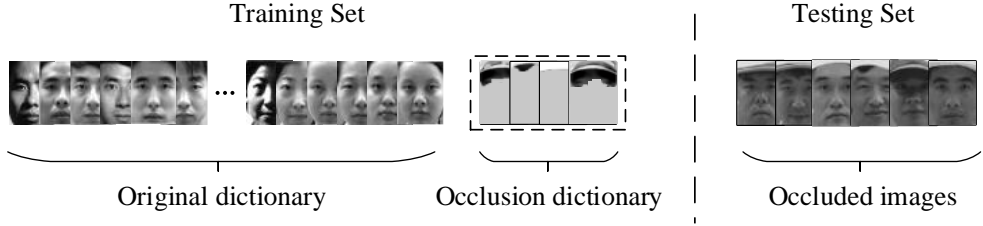

Fig. 12 Diagram of face recognition task with occlusion (hat) on CAS-PEAL database

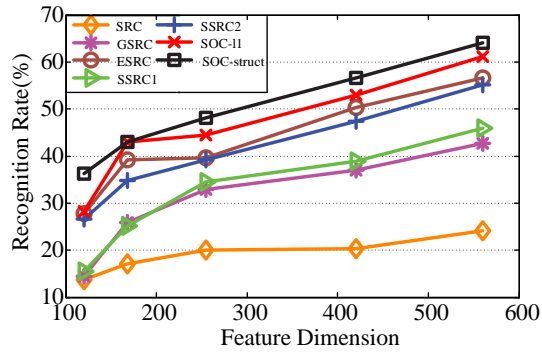

Fig. 13 The recognition rate on CAS-PEAL database.

In this challenging database, we consider the hat as a large and spatially continuous occlusion since it not only covers the hair but also changes the illumination over half face. As usual, dense solution caused by large occlusion results in poor classification in SRC. GSRC shows more distinct robustness than SRC thanks to its local characteristic. With the category-specific occlusion dictionary, ESRC, SSRC1 and SSRC2 are able to successfully handle the occlusion problem. The occluded image can be sparsely represented, enabling its label to be revealed by the sparse coefficients. However, they still suffer from the inexact representation incurred by imprecise occlusion samples. SOC achieves significant improvement compared with the state-of-the-art approaches, mainly because we exclude much non-occluded information out of occlusion mask. Purer occlusion samples naturally leads to better representation for testing images. Specifically, our algorithm outperforms SSRC1 by 9.63%, 8.15%, 8.89%, 9.26% and 8.89% in 120D, 168D, 255D, 420D and 560D. Additionally, structured sparsity also brings 8.15% improvement compared to $l_1$ sparsity in 120D.

*4.4. Robust face recognition with unlabeled occluded images*

As shown in 4.3, the proposed algorithm is effective in handling real occlusion with a well-trained occlusion dictionary. But in practical condition, we have to address in occlusion problem with more restricted precondition. In this section, all the labels in occluded images are erased. We are going to learn a more generic occlusion dictionary by algorithm 1, with unlabeled occluded training images.

*4.4.1 The AR database*

In the AR database, we randomly select faces from 10 random subjects, with each subject containing 30 images occluded by sunglasses and 30 images occluded by scarf, and their identities are erased. The approaches of collecting occlusion samples are descripted in 4.2. We also carry out testing on the four scenarios in 4.3.1, sharing the same training and testing images. Fig. 14 shows the experimental results. Note that SRC and GSRC do not rely on occluded training images and perform in the same way in 4.3.1. Hence we exclude them in this experiment.

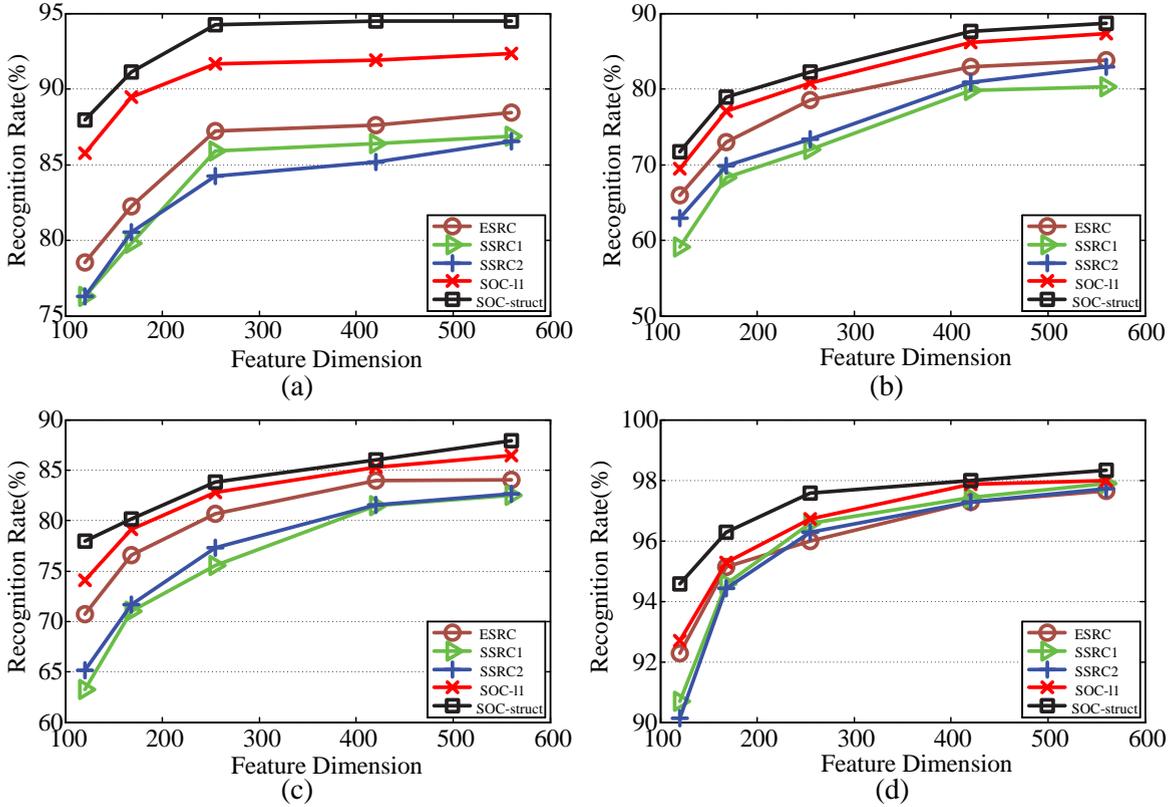

Fig. 14 The recognition rate in four scenarios on AR database. (a) Sunglasses images. (b) Scarf images. (c) Sunglasses and scarf images (d) Normal images

In this more restricted but common scenario, ESRC, SSRC1 and SSRC2 recognition rates are slightly lower than the results in section 4.3.1 because the corresponding sub-dictionary is not available. By subtracting the occluded image from the centroid of $D$ or projecting the occluded image on a subspace of another sub-dictionary, it greatly increases the noise in occlusion samples. Specifically, the accuracy of SSRC1 declines by 4.07%, 4.08 and 5.78% in occluded scenarios. However, SOC shows more robustness and achieves comparable results as in 4.3.1, even without labeled occluded images. The highest recognition rates in these four occluding scenarios are 94.50%, 88.67%, 87.93% and 98.34% in 560D, achieved by structured sparsity constraint. In low dimension (120D), we also obtain 11.66%, 12.60%, 14.72% and 3.9% improvement compared to SSRC1. The experiment demonstrates our previous claims. The face subspace, built by locality constrained dictionary, can nearly take the place of the original one to estimate the occlusion mask. Most importantly, it provides a more convenient and generic approach for us to collect occlusion sample in practical condition.

*4.4.2 The CAS-PEAL database*

In the CAS-PEAL database, the original dictionary consists of 1000 images. We randomly divide the images with hat into two groups. The first group containing 10 subjects (30 images per subject) is used to construct occlusion samples set. The size of occlusion dictionary are set to 20. We test on the second group.

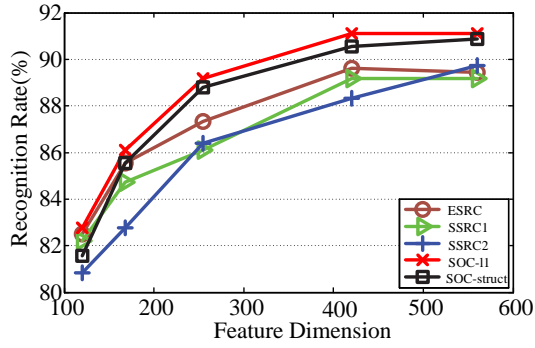

Fig. 15 The recognition rate on CAS-PEAL database.

In this challenging scenario, SOC successfully estimates the exact occlusion pattern, which indicates that the locality constrained dictionary can provide an alternative feasible subspace for testing images. It is useful in practical scenario since we can obtain more occlusion samples. Specifically, we improve the accuracy roughly from 8% to 10%. In these large occlusion scenarios, SOC achieves the best recognition accuracy among several competitive algorithms.

### 4.5. Robust face recognition with exaggerated expression

In practical scenario, some expressions such as yawning, laughing or screaming, may violate the original representation model as well. In our framework, we treat them as another kind of occlusion, arguing that a dictionary consisted of various expressions can be learned from the training images in advance for stronger robust.

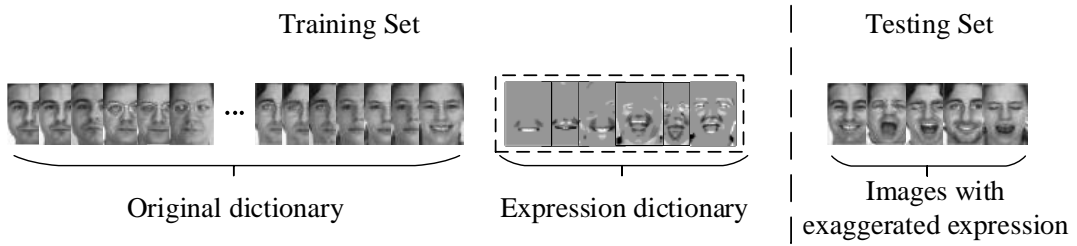

Fig. 16 Diagram of face recognition task with exaggerated expression on AR database

In AR database, we only consider the images without disguise. We manually choose 10 images with faint expressions per subject for training. The remaining 4 images with exaggerated expressions are divided into training samples of 10 subjects and test samples of 90 subjects. We learn an expression dictionary of 30 atoms from 40 expression samples. We have 999 face atoms and 30 expression atoms in dictionary and 360 testing images. The recognition rates of ESRC, SSRC1, SSRC2 and SOC are given in Fig. 17.

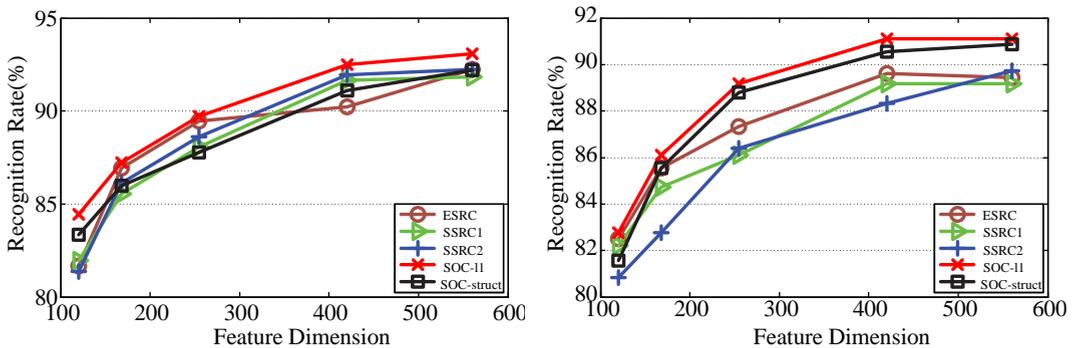

Fig. 17 Expression recognition rate on AR database. (a) Recognition rate with labeled occluded training images. (b) Recognition rate with unlabeled occluded training images.

One can see that various expressions can be treated as a kind of generalized occlusions, since they merely appear around the mouth, eyes and cheeks. Compared with other category-specific occlusion dictionary, our method still brings significant improvement, especially training with unlabeled occlude. It is notable that $l_1$ sparse constraint outperforms structured sparse constraint. The main reason is that we group all the expression atoms in a sub-dictionary, causing much mutual coherence when representing.

*4.6. Running time*

In this section, we compare the running time of the five algorithms, including SRC, GSRC, ESRC, SSRC1, SSRC2 and SOC. We record the elapsed time of occlusion pattern collecting, occlusion dictionary training and recognizing, in which we learning an occlusion dictionary of 30 atoms from 60 occlusion samples, in resolution of $83 \times 60$. Differently, GSRC needs to compute Gabor feature of an identity matrix with the size of $4980 \times 4980$, and compact it to a Gabor occlusion dictionary with the size of $4980 \times 30$, which cost a long time. When it comes to the recognition stage, we down-sample the dimension to 560. Other experimental settings are the same as in section 4.3.

Table 1. The average running time (second) of several algorithms (*None* means the time is negligible)

| Method | Occlusion sample collecting (1 occlusion image) | Occlusion dictionary training | Recognizing (1 test image) |
| --- | --- | --- | --- |
| SRC | None | None | 4.13 |
| GSRC | 0.03 | 1765.23 | 2.66 |
| ESRC | 0.0021 | 1.08 | 2.59 |
| SSRC1 | 0.17 | 1.10 | 2.56 |
| SSRC2 | 0.17 | 321.56 | 2.62 |
| SOC-$l_1$ | 0.28 | 1.06 | 2.60 |
| SOC-structured | 0.28 | 1.07 | 2.72 |

*4.7. Occlusion recognition*

As one can see, each atom in occlusion dictionary is related to a specific category of practical occlusion. Thus the type of occlusion is also able to be recognized as the face does. We conduct a simple experiment with the image resolution of $17 \times 15$. 700 normal images and 100 (50 sunglasses and 50 scarf) occlusion samples constitutes the compound dictionary. 500 images with sunglasses and 500 images with scarf are used for testing. Since the occlusions only includes sunglasses and scarf, which degenerates to a binary classification problem. SOC obtains a perfect classification result.

Table 2. The recognition rate of occlusion

| Method | Recognition rate (%) |
| --- | --- |
| SOC - $l_1$ sparsity | 100.00 |
| SOC - Structured sparsity | 100.00 |

*4.8. Rejecting invalid test images*

In a practical face recognition system, it is also important to reject the outlier subject of face and occlusion which are not included in training set. The rejection rule is based on RDI. In AR database, we randomly choose 50 subjects including 7 clean images and 2 occluded images (sunglasses and scarf) for training, and the remaining are used for testing. So we has 849 valid images, and 1300 invalid images should be rejected.

On the other hand, we also test the performance of rejecting unknown occlusions. In our experiment, sunglasses and scarf are regarded as valid occlusion. We also manually generate images with unknown occlusion by covering an unrelated image on random region. The occlusion size varies from 10% to 50%. We construct the dictionary in the same way with 4.2, but choose 1000 images with sunglasses and scarf as valid test images, and cover the rest normal images as invalid ones. All the images are resized to $17 \times 15$. Fig. 18 gives the receiver operating characteristic (ROC) curve for face and occlusion. We can see both the $l_1$ sparsity and structured sparsity perform well in rejecting invalid images.

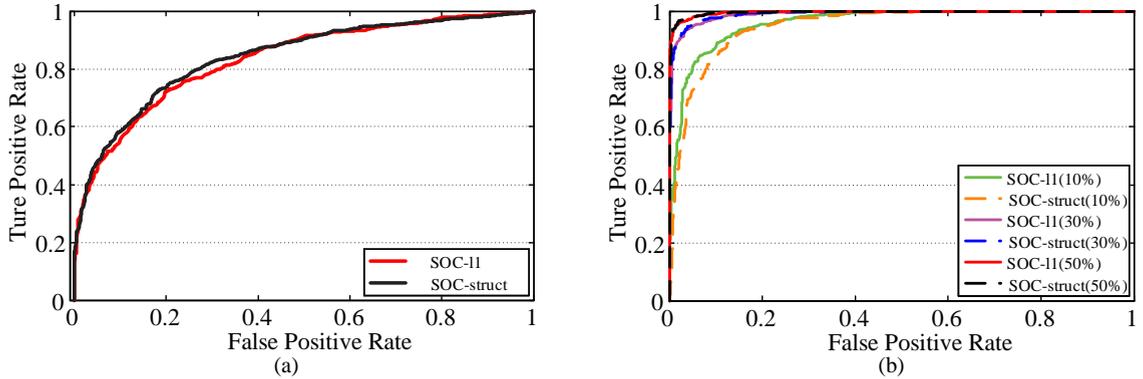

Fig. 18 (a) ROC curves for invalid test image rejection. (b) ROC curves for unknown occlusion rejection

*4.9. The effect of varying size of occlusion dictionary*

The occlusion dictionary is crucial in the sparse representation stage. A reasonable occlusion set should include sufficient training samples for representation. In this section, we exploit the relationship between the recognition performance and the size of occlusion dictionary. The scenario of both sunglasses and scarf is presented. The images are resized to the resolution of $17 \times 15$. The size of original dictionary is 700, with 7 faces each subject. We vary the size of occlusion dictionary to 2, 3, 5, 7, 10, 20, 30, 40, 50 and 60, plotting the results in Fig. 19.

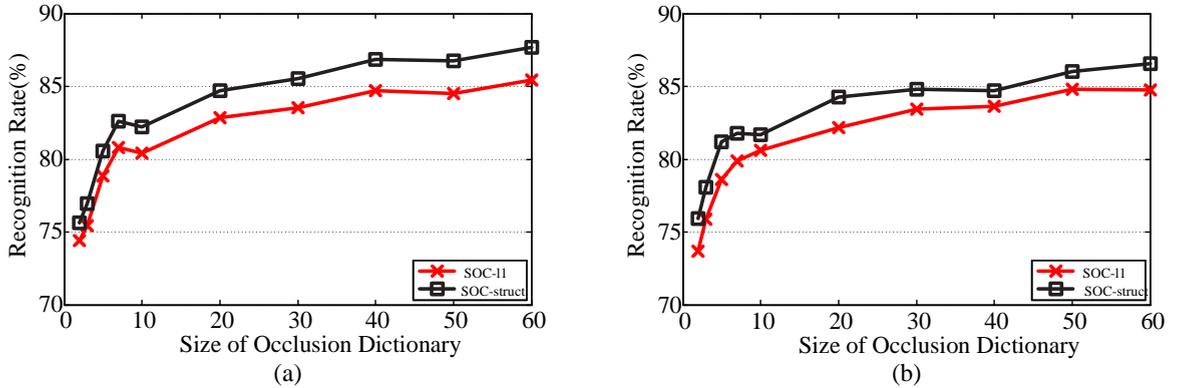

Fig. 19 The relationship between the recognition rate and the size of occlusion dictionary. (a) Construct the occlusion dictionary using labeled occluded images. (b) Construct the occlusion dictionary using unlabeled images.

From Fig. 19, it can be learned that a redundant occlusion dictionary is not necessary, since redundant occlusion atoms contributes little to representation and classification. For higher efficiency, we can design a compact occlusion dictionary for practical use, which can also guarantee a satisfactory result. This experiment further illustrates that LCD serves a similar role to the corresponding sub-dictionaries.

## 5. Conclusion and Future Work

In this paper, we propose a classification framework, which simultaneously separates the occlusion and classifies the test image by coding over the occlusion dictionary with a structured sparsity constraint. By taking advantage of the occlusion mask, the occlusion dictionary is more generic and contains less noise, enabling itself to represent occluded images more precisely. From a practical standpoint, we propose an approach to estimate occlusion mask, achieving comparable estimate result under a more restricted condition. In this approach, we demonstrate that the locality constrained dictionary serves an alternative subspace for test images. For stronger robustness, structured sparsity is used in the proposed algorithm. Moreover, in category-specific occlusion dictionary, the occlusion also becomes recognizable. In addition, a new rejection rule called RDI is proposed for verifying the validity of face and occlusion. Comprehensive experimental results show that SOC can better deal with face recognition with occlusion than most existing well-performing algorithms.

Further improvements of SOC can be made. For example, kernelization [32] can be performed in SOC. More robust coding technique [33] can be combined with SOC. In order to design a more robust system for practical scenarios, we are also interested in performing alignment via dictionary based methods, e.g., by incorporating a deformation dictionary into our framework. It remains future explorations.

## Acknowledgements

This work is supported by Guangzhou Science and technology Project (Grant No. 2014J4100247) and NSFC (Grant No. 61471174).